\documentclass{article}

\usepackage{arxiv}

\usepackage[utf8]{inputenc} 
\usepackage[T1]{fontenc}    
\usepackage{hyperref}       
\usepackage{url}            
\usepackage{booktabs}       
\usepackage{amsfonts}       
\usepackage{nicefrac}       
\usepackage{microtype}      
\usepackage{lipsum}		
\usepackage{graphicx}
\usepackage[authoryear]{natbib}
\usepackage{doi}
\usepackage{caption}
\usepackage{threeparttable}

\title{Are Large Language Models the future crowd workers of Linguistics?}

\date{January 31, 2025}	

\author{
    \href{https://orcid.org/0009-0003-9860-5437}{\includegraphics[scale=0.06]{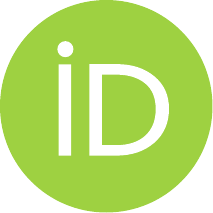}\hspace{1mm}Iris~Ferrazzo} \\ %
    Bonn Center for Digital Humanities (BCDH); Romance Languages Department \\
    Universität Bonn \\
    \texttt{iris.ferrazzo@uni-bonn.de}
}

\renewcommand{\shorttitle}{\textit{Are Large Language Models the future crowd workers of Linguistics?}}

\hypersetup{
pdftitle={Are Large Language Models the future crowd workers of Linguistics?},
pdfsubject={q-bio.NC, q-bio.QM},
pdfauthor={Iris~Ferrazzo},
pdfkeywords={large language models, crowdsourcing,
romance linguistics, computational linguistics, nlp},
}

\begin{document}
\maketitle
\begin{abstract}
Data elicitation from human participants is one of the core data collection strategies used in empirical linguistic research. The amount of participants in such studies may vary considerably, ranging from a handful to crowdsourcing dimensions. Even if they provide resourceful extensive data, both of these settings come alongside many disadvantages, such as low control of participants' attention during task completion, precarious working conditions in crowdsourcing environments, and time-consuming experimental designs.

For these reasons, this research aims to answer the question of whether Large Language Models (LLMs) may overcome those obstacles if included in empirical linguistic pipelines. Two reproduction case studies are conducted to gain clarity into this matter: \citet{cruz2023} and \citet{lombard2021}.
The two forced elicitation tasks, originally designed for human participants, are reproduced in the proposed framework with the help of OpenAI's GPT-4o-mini model. Its performance with our zero-shot prompting baseline shows the effectiveness and high versatility of LLMs, that tend to outperform human informants in linguistic tasks. The findings of the second replication further highlight the need to explore additional prompting techniques, such as Chain-of-Thought (CoT) prompting, which, in a second follow-up experiment, demonstrates higher alignment to human performance on both critical and filler items. Given the limited scale of this study, it is worthwhile to further explore the performance of LLMs in empirical Linguistics and in other future applications in the humanities.
\end{abstract}


\section{Introduction}
Empirical linguistic studies are often based on evidence gathered through tasks completed by humans. This data collection strategy allows for a richer view on how people perceive, judge, and use language by relying on the multitude-perspective of the crowd \citep{ahmadabadi_2024}. This applies to studies involving groups of informants of varying sizes, including those recruited via crowdsourcing platforms such as Prolific\footnote{https://www.prolific.com/} and Amazon Mechanical Turk\footnote{https://www.mturk.com/}. Research suggests that crowd workers are a viable resource for exploring a variety of empirical linguistic phenomena, including syntactic properties of minority languages \citep{sheehan_crowdsourcing_2019}, language proficiency judgments \citep{thwaites_crowdsourced_2024}, lexical diversity \citep{khalilia_crowdsourcing_2024}, and pragmatic implicatures of conversational negation \citep{capuano_semantic_2021}, among others. 

However, this comes with a cost: humans are demanding, expensive, and difficult to engage in reproducible research settings. Although quick and extensive data collection is made possible through crowdsourcing by reaching matching subscribers with newly published studies, it is still questionable how researchers can actively control the quality and the attention of their participants' performance.

Concerns over the precarity of the working conditions in crowdsourcing have also been broadly voiced \citep{van_zoonen_algorithmic_2024}. Furthermore, with the widespread of computational models, especially Large Language Models (LLMs), that can perform human-like tasks and even outperform the accuracy of crowd workers \citep{kocon_chatgpt_2023, gilardi_chatgpt_2023} and experts \citep{tornberg_chatgpt-4_2023, openai_gpt-4_2023}, illicit evidence of their usage in crowdsourcing studies is currently under observation \citep{veselovsky_artificial_2023}. Questions both in Linguistics and in Natural Language Processing (NLP) and Generation (NLG) arise around textual data collected via crowdsourcing that may be already produced by machines, rather than by the hired human crowd workers. In order to overcome the shortcomings and costs of managing human participants in such empirical studies, the present paper investigates to what extent researchers of humanities, in this case of empirical Linguistics, may use LLMs in their research workflows as a replacement of, or in addition to, human informants and crowd workers.

While research on the analytic abilities of LLMs as task solvers, especially using ChatGPT, GPT-4, LLama, Gemini, and Claude, is a current NLP matter and shows the potential and limitations of these models across different tasks and disciplines \citep{kocon_chatgpt_2023}, less attention is given to the possible operationalization of these models in humanities and social sciences research. Furthermore, this line of research lies in the controversy of the replacement of humans through machines, which is perceived as critical \citep{ziems_can_2024}. 
Filling these gaps, this paper aims to:
\begin{itemize}
  \item address the question of whether LLMs can become the next generation of crowd workers for empirical linguistic studies. Linguistic research conducted with human participants usually goes beyond mere annotation tasks (see common NLP-tasks such as sentiment analysis, Part-of-Speech-tagging, syntactic parsing, text classification, among others) and grounds in language processing, judgement, and generation under very specific analytical conditions,
  \item extend the range of action of LLMs in the humanities by testing their behaviour on task outlines that were designed traditionally for humans,
  \item develop an approachable basic prompt engineering framework that can be easily reused also by non-programmers who want to explore LLMs' potential in their research, 
  \item support the line of work that intends to overcome the English-centric training and usage of LLMs by studying their performance on Romance languages.
\end{itemize}

These objectives are addressed in the present paper by replicating two empirical linguistic studies that have been conducted on human participants. In the proposed pipeline, their work is replaced by LLMs, specifically OpenAI's GPT-4o-mini, here used to reach crowd workers-like performance levels. Although this study cannot cover the entire spectrum of areas of interest in linguistic research, the results point in the direction of an alignment between LLMs and human participants that may support a needed interdisciplinary opening between NLP and the humanities. In summary, we report the following findings:

\begin{enumerate}
    \item The proposed pipeline demonstrates adaptability and applicability to a wide range of instruction-based tasks, underscoring the broader significance of this research.
    \item In the first replicated study, GPT-4o-mini's performance aligns well with human participants, highlighting LLMs' potential in empirical linguistic research.
    \item The second task reproduced introduces a more nuanced perspective on LLMs-to-humans alignments and offers an opportunity to explore strategies for managing outcomes, especially through the use of specific prompting techniques (see Chain-of-Thought prompting).
    \item GPT-4o-mini outperforms human informants in all experimental conditions tested.
\end{enumerate}
The paper is structured as follows: first, it briefly reviews the state of the art of LLMs' performance in  instruction-based tasks in NLP and in empirical Linguistics (Section ~\ref{sec:related} ), second, it outlines the replication methodology designed for our two case studies, including details about the model and prompt engineering (Section ~\ref{sec:method} ). Third, the results of the two replications are presented (Section ~\ref{sec:repli1}  and ~\ref{sec:repli2}) and discussed in light of the aforementioned research questions (Section ~\ref{sec:discu}). Finally, we conclude the paper with a brief summary of newly gained insights, research limitations, and future work directions (Section ~\ref{sec:conclu}).

\newpage
\section{Related Work}
\label{sec:related}
\subsection{Large Language Models as crowd workers in NLP}

LLMs are remarkable data annotators. Their pre-training and supervised instruction fine-tuning build intelligent assistants that mimic human behaviour \citep{openai_gpt-4_2023}. Consequently, LLMs exhibit exceptional capabilities in adhering to instructions designed to elicit specific model responses, while offering a user-friendly interaction experience. Within the field of NLP, ongoing research investigates these abilities through the development of frameworks that either incorporate human informants, or that operate independently of them \citep{kocon_chatgpt_2023}. Even if LLMs do not consistently outperform humans across all tasks, they are well-positioned to make meaningful contributions to social science analysis in collaboration with human researchers \citep{ziems_can_2024}. These pipelines often task computational models with performing direct or indirect annotation \citep{van_dalfsen_direct_2024}, using either supervised or unsupervised methods \citep{horych_promises_2024}, or by establishing semi-automated annotation workflows \citep{ostyakova_chatgpt_2023, xu_role_2024}. Recent findings suggest that LLMs match or surpass human crowd workers in numerous NLP tasks \citep{tornberg_chatgpt-4_2023, ziems_can_2024, he_annollm_2023, kocon_chatgpt_2023, gilardi_chatgpt_2023}. However, these studies predominantly focus on crowdsourcing scenarios where participants are required to assign labels to data, rather than engage in language generation or exercise complex judgment. \citet{wu_llms_2025} use LLMs in the replication of previous crowdsourcing pipelines in a course assignment and describe high variance in the results according to task complexity. They suggest that analysing LLMs within established pipelines or workflows provides a clearer insight into their strengths and limitations; accordingly, this paper aligns with that objective.

\subsection{Large Language Models as crowd workers in empirical Linguistics}

Although disciplines outside NLP, such as cognitive research \citep{dillion_can_2023} and economics \citep{gui_challenge_2023}, have already explored whether LLMs may replace or at least productively simulate human participants in empirical studies, this matter remains understudied in Linguistics. \citep{Argyle_2023} provided initial evidence suggesting that language models can serve as proxies for specific human sub-populations, offering new avenues for research in the social sciences. This has been applied to linguistic tasks such as sentiment analysis \citep{borst_death_2023, rebora_comparing_2023}, topic detection \citep{kosar_comparative_2024}, semantic annotation \citep{gilardi_chatgpt_2023}, paraphrase generation \citep{cegin_chatgpt_2023}, and text classification \citep{de_lange_benchmarking_2024}, but has not been extended systematically outside NLP-related annotation tasks, apart from few exceptions (see \citep{karjus2024machineassisted} on a set of case studies such as annotation of semantic change detection, and \citep{Karjus2024-zg} on measurements of linguistic divergence in US American English across political spectra). Even with a linguistic core interest, these annotation tasks display a strong connection to NLP-based pipelines rather than to actual ongoing linguistic research. Linguistics researchers continue to focus mostly on experimental designs involving human participants, or they engage with LLMs primarily through the ChatGPT interface \citep{dynel_lessons_2023, han_chatgpt_2024, duncan_does_2024}, rather than exploring a broader range of models within actual coding environments. Empirical linguistic studies vary in design, employing diverse data collection methods, analytical goals, and participant criteria, offering a valuable opportunity to evaluate LLMs as crowd workers.

\section{Methods}
\label{sec:method}
\begin{figure*}[!t]
\centering
\includegraphics[width=1\linewidth]{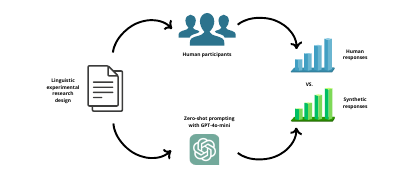}
\caption{Illustration of our approach for quantifying the alignment between human participants' and GPT-4o-mini's responses in empirical research designs with linguistic focus.}
\label{figure:approach}
\label{fig_wide}
\end{figure*}

The overall approach to test LLMs as linguistic crowd workers is illustrated in Figure~\ref{figure:approach} and is described in detail below, including chosen models, replication procedure, and materials used.  

\subsection{GPT-4-mini as crowd worker}
A closed-source model of the OpenAI family, GPT-4o-mini, is chosen to assess LLMs in the role of crowd workers in the proposed study replications. Although closed-source LLMs can be challenging in matters of limited transparency of their training data and of additional costs of their API usage, these models deliver state-of-the-art results and offer a successful user experience. This facilitates a beginner-friendly introduction to these models, making them accessible to humanities researchers who wish to experiment with the models' API using a lightweight code base without requiring access to high-performance computing infrastructures. GPT-4o-mini is selected for the present investigation as OpenAI's most cost-efficient small model to the present moment. Our baseline approach involves zero-shot prompting of GPT-4o-mini. The model is presented with a prompt that includes information about the participants' profile to be impersonated by the model, along with the task instructions from the studies being replicated. A recurrent meaningful piece of information about the informants is their mother tongue. Other prompt engineering techniques were discarded to first establish a baseline that evaluates the model's responses in a way that is comparable with the original human-based study.  In this paper, we build this initial baseline by prompting the model directly without fine-tuning. Although fine-tuning has been shown to enhance LLMs' performance in text annotation \citep{alizadeh2025-cm}, this paper aims to evaluate the models in an accessible coding environment while ensuring computational efficiency and assessing their inherent abilities.
An overview of the prompts used in the different experiments can be found in Appendix~\ref{appendix:C}.

\subsection{Replicating empirical linguistic studies with LLMs}
The goal of the paper is to replicate the workflow of two studies that were conducted in the field of empirical Linguistics, this time by replacing human participants with LLMs as crowd workers. We test the performance of LLMs in linguistic tasks that go beyond mere annotation and address the question of model-to-human judgment alignment. The studies we choose to replicate with LLMs match a set of fixed parameters in order to fill in the previously mentioned research gaps: 
\parbox{\linewidth}{
\begin{itemize}
  \item The analytical focus must be set on Romance languages,
  \item The task designed for their human participants is based on language generation and/or understanding, not just labelling, 
  \item The research question is rooted in linguistics rather than originating from the field of NLP.
\end{itemize}
}
Two forced-choice binary elicitation tasks that match the aforementioned criteria were selected. They involve Spanish and English code-switching environments \citep {cruz2023} and neological intuition in French native speakers \citep {lombard2021}. Both studies prompt responses from human participants on a computer screen, but they elicit them in two different forms: a discourse completion \citep {cruz2023} conducted in person and a linguistic judgment survey (by pressing "yes" or "no" buttons; Lombard, Huyghe, and Gygax \citeyear{lombard2021}\citep{lombard2021}) completed on the \textit{Qualtrics} online platform. For the sake of clarity and consistency throughout the paper, the two studies will be referred to as Cruz\_23 and Lombard\_21.

In our replications we keep the experimental design as close as possible to the original. The main contribution of our research is to adapt their pipeline to LLMs as crowd workers. For this reason, it does not intend to offer a better or different version of the original experiments and no judgement is passed about the quality of their empirical pipeline.

\subsection{Replication pipeline}
\begin{figure}[hbt!]
\centering
\includegraphics[width=0.8\linewidth, trim=0 20 90 0, clip]{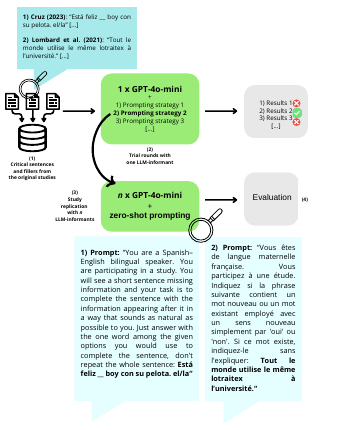} 
\caption{Replication pipeline.}
\label{figure:replipipeline}
\end{figure}

We replicate Cruz\_23 and Lombard\_21 through the same straightforward coding framework, which can be accessed online (see Appendix~\ref{appendix:A}). This shows the versatility of the proposed methodology that can be used to replicate several new crowdsourcing studies and to approach prompting techniques to test LLMs in different research settings. The code pipeline is divided into four main parts (see \autoref{figure:replipipeline}): (1) first, the critical stimuli and the distractor sentences are read and pre-processed for further analysis. (2) Then, different prompting strategies are operationalised on only one "LLM-informant" at the time, i.e. in one query using GPT-4o-mini. This keeps the API usage costs under control and gives users the possibility to try different prompt wordings until the expected model's output is matched. As previously mentioned, zero-shot prompting is chosen as main prompting strategy for our replication after first pilot studies. The intermediate trials with one "informant" at the time can be locally saved and cited as a reference to justify further steps of the pipeline. (3) Next, code is provided to extend the experiment to more "LLM-informants", and the best-performing prompting strategy, tested in the previous step, is repeated as many times as we find human participants in the original paper to achieve full replication. Since the model is prompted in different runs that are not linked to one another, the answers are unrelated between "LLM-informants" and contamination is prevented. This would not be the case in the ChatGPT interface, where model responses to the same prompt are generated in one single query. This method simulates the performance of several human informants who work singularly on their task. In other words, the number of participants of the original linguistic study is used to define the number of prompting runs. (4) In the end, the synthetic answers are collected and saved locally to evaluate them against the human baseline of the original paper. The results are then discussed quantitatively and qualitatively. The pipeline is repeated twice per replicated study to ensure cross-validation and statistical relevance of the results. Mean values across the two runs are reported in the following sections.

\section{Study 1: Modulating gender assignment in Spanish–English bilingual speech (Cruz 2023)}
\label{sec:repli1}

\subsection{Study overview}
We replicate the task originally devised by Cruz\_23, whose goal was to 
examine linguistic factors (i.e., semantic a.k.a. biological gender, analogical gender, and other-language phonemic cues, see Table~\ref{table:conditionsCruz}) that may affect gender assignment (masculine or feminine) to English nouns occurring in code-switches. As part of their experiment, crowd workers completed first a forced choice elicitation task involving a code-switching environment: switches between Spanish determiners and English nouns (e.g., \textit{Ya estamos en \_\_ plane}, in English: We are already in \_\_ plane, see Cruz\_23: 586), where participants had the binary choice between the two possible Spanish determiners: \textit{el} (masculine) and \textit{la} (feminine). This is linguistically relevant since Spanish and English exhibit different distributional patterns in terms of gender assignment: Spanish employs a binary gender system, allocating every noun to either the masculine or feminine category, whereas the grammatical gender of English is absolute. Accordingly, Cruz\_23 investigates whether the gender of the Spanish translation equivalent (analogical gender) influences the assignment mechanism used in code-switched speech or whether this mechanism functions independently of the gender system of the other language.

The experimental setting of task 1 comprehended 34 participants, early bilinguals, who reacted to 80 sentences from the CESA Corpus \citep{carvalho_corpus_2012} containing critical stimuli (English nouns equally divided into human-denoting nouns and inanimate nouns) and 75 distractor nouns (with binary answer options that do not display any marks for gender assignment in Spanish, i.e. \textit{mi} and \textit{su}, in English: my and his/her/its). The experimental stimuli can be found in Cruz\_23's \href{https://www.cambridge.org/core/journals/bilingualism-language-and-cognition/article/linguistic-factors-modulating-gender-assignment-in-spanishenglish-bilingual-speech/CB7FD8DBFAF5EE5E4957F72B651D8C6A#supplementary-materials}{supplementary materials}.  Table~\ref{table:conditionsCruz} gives an overview of the eight tested conditions according to semantic and/or grammatical gender, animacy of the referent, and phonemic cues about the grammatical gender of the Spanish translation. 

\begin{table}[hbt!]
\centering
\vspace{5pt} 
\captionsetup{justification=justified, singlelinecheck=false} 
\caption{Conditions tested in Cruz\_23, Task 1} 
\label{table:conditionsCruz}
\begin{threeparttable} 
\begin{tabular}{llll} 
\toprule
\textbf{Condition} & \textbf{Variables} & \textbf{Example} & \textbf{n=80} \\ 
\midrule
1 & M\tnote{*}, animate, strong cue\tnote{**} & Nephew (\textit{sobrino}) & n=10 \\
2 & M, animate, no cue\tnote{**} & King (\textit{rey}) & n=10 \\
3 & M, inanimate, strong cue & Roof (\textit{techo}) & n=10 \\
4 & M, inanimate, no cue & Pencil (\textit{lápiz}) & n=10 \\
5 & F\tnote{*}, animate, strong cue & Niece (\textit{sobrina}) & n=10 \\
6 & F, animate, no cue & Actress (\textit{actriz}) & n=10 \\
7 & F, inanimate, strong cue & Movie (\textit{película}) & n=10 \\
8 & F, inanimate, no cue & Salt (\textit{sal}) & n=10 \\ 
\bottomrule
\end{tabular}
\begin{tablenotes}
    \item[*] \textit{M }stands for masculine, \textit{F} stands for feminine.
    \item[**] \textit{Strong} vs. \textit{no cue} variables refer respectively to whether grammatical gender information is provided through the morphology of the Spanish translation of the English target noun or not.
\end{tablenotes}
\end{threeparttable}
\end{table}

\subsection{Results and error analysis}
\begin{figure}[hbt!]
\centering
\includegraphics[width=0.6\linewidth, trim=0 35 0 25, clip]{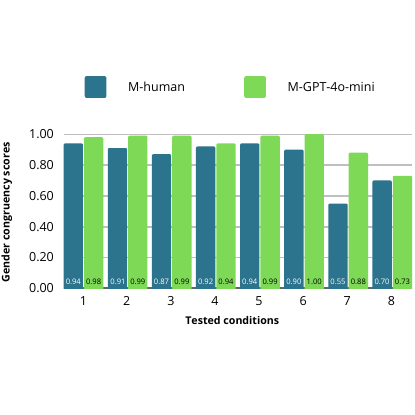} 
\caption{Comparison of mean responses (M) in terms of gender congruency between human participants and GPT-4o-mini.}
\label{figure:Cruz}
\label{fig_sim}
\end{figure}

Cruz\_23 describes their findings in terms of gender congruency, defined as the proportion of Spanish determiners chosen by their study participants whose grammatical gender aligns (gender congruent) or does not align (gender incongruent) with the Spanish translation of the English target noun. Their descriptives and our corresponding findings are shown in Figure~\ref{figure:Cruz} for direct comparison. Across the eight tested conditions, GPT-4o-mini shows comparable performance tendencies to human participants and reveals gender congruency patterns.

As a general trend, GPT-4o-mini exhibits higher gender congruency choices in all experimental conditions compared to human participants. Strong alignments with human choices are found with human denoting nouns (Conditions 1, 2, 5, 6) that are assigned by the model a determiner of the grammatical gender in line with the referent's semantic (a.k.a. biological) gender almost without doubt (Conditions 1, 2, 5, and 6 combined; M=0.99). As in the original study, proportions of gender-congruent selections for inanimate nouns were higher for masculine gender (Conditions 3 and 4 combined; M-human=0.89, M-GPT=0.96) compared to feminine gender (Conditions 7 and 8 combined; M-human=0.62, M-GPT=0.80). English target nouns whose Spanish translation exhibits morphological cues of feminine grammatical gender (e.g., \textit{border} - in Spanish: \textit{fronter\textbf{a}}. Condition 7; M-human=0.55, M-GPT=0.88) are assigned gender congruent determiners more often than without cues for grammatical gender (e.g., \textit{honey} - in Spanish: \textit{miel}. Condition 8; M-human=0.70, M-GPT=0.73). In Condition 7, the model significantly outperforms human participants, whereas in Condition 8, both humans and LLMs achieve comparably lower gender congruency levels. The gender incongruent model's decisions and their corresponding scores can be found in Appendix~\ref{appendix:B}.

The task duration of our reproduction with GPT-4o-mini was approximately 72 seconds for each iteration through all the critical and distractors items, which stands for one informant's performance in the experimental design. The study duration reported with human participants is 50 minutes per person, which includes a first verbal exchange between the researcher and the participant, followed by the completion of tasks 1 and 2. Since we replicate only task 1, we can expect it to take originally approximately 25 minutes per participant.

\section{Study 2: Neological intuition in French (Lombard et al. 2021)}
\label{sec:repli2}

\subsection{Study overview}
As second approach to our purpose, we replicate Lombard\_21 metalinguistic task of novel word identification with French (non-linguist) native speakers. To assess what classes of linguistic information about neologisms may help native speakers to recognise their occurrence in a given speech sample, two variables are analysed: formal novelty and lexical regularity. The main hypothesis is that informants may tend to spot neologisms easier when they display high levels of formal novelty (i.e., they are newly created words rather than existing words used with new meanings) and of lexical irregularity (i.e., they are created through irregular rather than regular word formation processes). Morphological neologisms are new words created through modifications to the form and, typically, the meaning of a lexical base (1), whereas semantic neologisms are existing words that acquire new, additional meanings (2):

    \begin{enumerate}
        \item e.g., \textit{Certaines célébrités vont à contre-courant, en se faisant \textbf{détatouer}} (Lombard\_21: 4, in English: Some celebrities go against the tide by having their tattoos removed)
        \item e.g., \textit{Dans une relation \textbf{toxique}, les tensions et les critiques sont omniprésentes} (Lombard\_21: 4, in English: In a toxic relationship, tension and criticism are omnipresent)
    \end{enumerate}
    
Participants (n=68) were instructed to first read a set of French sentences and then judge whether they contain neologisms. In a positive case, they should identify novel words by clicking on them and their reaction times were logged. The stimuli comprehended 80 experimental sentences including neologisms that were created ad hoc for the study purpose, and 40 filler sentences without neologisms. They can be accessed in Lombard\_21's \href{https://ars.els-cdn.com/content/image/1-s2.0-S0024384121000279-mmc1.pdf}{supplementary materials}. Each experimental condition included 20 stimuli (see Table~\ref{table:conditionsLombard}). The critical sentences had neologisms (nouns and verbs) placed in different syntactic positions to prevent the formation of repeated decision patterns, displayed a standardized length, and presented syntactic simplicity. Filler sentences were constructed analogously, they only differ in the absence of neologisms.

\begin{table}[hbt!]
\centering
\caption{Conditions tested in Lombard\_21.}
\vspace{5pt}
\label{table:conditionsLombard}
\small 
\begin{tabular}{llll}
\toprule
\textbf{Condition} & \textbf{Variables} & \textbf{Example} & \textbf{n=80} \\ 
\midrule
1 & Irreg. morpho. & \textit{Maigrimanger} (‘eat little’) & n=20 \\
2 & Reg. morpho. & \textit{Surcomplimenter} (‘overcompliment’) & n=20 \\
3 & Irreg. semantic & \textit{Jardiner} (‘do gardening’/‘ponder’) & n=20 \\
4 & Reg. semantic & \textit{Brise} (‘breeze’/‘small amount of events’) & n=20 \\
\bottomrule
\end{tabular}
\end{table}

\subsection{Results and error analysis}
The original study reports the average of 20 minutes registered by the human participants to complete the survey online. Our reproduction shows an average of 68 seconds per "LLM-informant".

\begin{figure}[hbt!]
\centering
\includegraphics[width=0.6\linewidth, trim=0 35 0 25, clip]{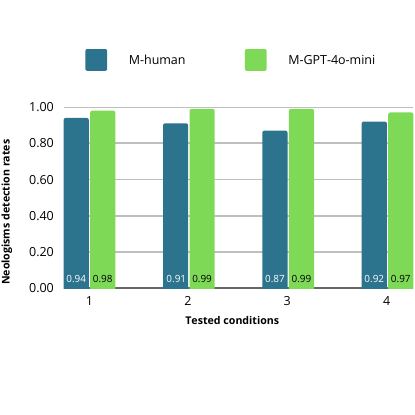} 
\caption{Comparison of mean responses (M) in terms of neologism detection between human participants and GPT-4o-mini.}
\label{figure:Lomb}
\label{fig_sim}
\end{figure}

Lombard\_21's discussion of their results is twofold: first, accuracy of neologism detection is measured, and second, participants' reaction times in correct detections are compared across the four experimental conditions. The first approach is illustrated in Figure~\ref{figure:Lomb}. LLM-informants of our replication outperform humans in the task of neologism detection. Accuracy is above ninety-seven percent (M-GPT-4o-mini=0.985) in all four conditions tested. This aligns with the overall findings of the original paper, as human participants also demonstrated high accuracy, albeit slightly lower than that of our LLM (M-human = 0.91). The analysis of reaction times recorded by GPT-4o-mini in all iterations through both critical and filler items indicates that this measurement does not serve as a reliable parameter to gain insights into the model’s reasoning process, as it does for human participants. Given that computation time is influenced by various external factors, such as hardware, model architecture, input complexity, and optimization techniques, it cannot be considered a meaningful evaluation parameter of the model's performance. Consequently, our results assessment focuses exclusively on neologisms detection rates.

A closer analysis of GPT-4o-mini's results in our replication shows one small drop in performance registered in Condition 4 (M-GPT-4o-mini=0.97), signalling that if the model experiences any difficulties in detecting neologisms, it is mostly when these words are not morphologically new, but are used with irregular new, extended meanings. The error analysis of the model's mistakes in detecting neologisms suggests that these are concentrated on a small number of individual words, with one word in particular (\textit{velours}, in English "velvet"/"velvet clothes", Condition 4) accounting for fifty percent of the errors in all conditions. This potentially indicates the word as an outlier. When this word is excluded from the total error count, the distribution of error scores across the tested conditions shifts, neutralising differences between conditions. GPT-4o-mini excels at the task and its performance is not influenced by linguistic parameters as it is the case with human participants. 

The most surprising aspect of the error analysis is not the error rates observed in the experimental conditions, but rather those associated with the fillers. In the original study, filler sentences correspond to items that do not contain neologisms, with the expected response from crowd workers being "no." Since this aspect is not addressed in the main paper, direct comparisons between human and LLM-informants are not possible. However, the model makes significantly more mistakes on filler items compared to critical stimuli, and reaches a detection accuracy of only 0.77. This outcome is unexpected, suggesting that the model tends to favour positive responses rather than simply answering with "no." A such behaviour can be potentially related to OpenAI models that are trained to act as helpful assistants, which could prioritize optimizing user experience over simply adhering to the task requirements of the study. For this reason, we attempt to adjust the model's performance through two follow-up experiments.

\subsubsection{Follow-up 1: Modification of the zero-shot prompting function}
First, we conduct an experiment and modify only one parameter of the replication pipeline: the zero-shot prompting function. In this function, the \texttt{role} parameter (i.e., {\texttt{"role": "system"}}) is used to define the role designed for the language model during task completion. OpenAI’s chat models, including GPT-4o-mini, follow a structured message format where each message has a \texttt{role}. The three main roles are:
\begin{enumerate}
    \item {\texttt{System}}: It provides high-level instructions that define the assistant's behaviour.
    \item {\texttt{User}}: It represents the user input or prompt.
    \item {\texttt{Assistant}}: It stands for previous model's response.

\end{enumerate}
In the first follow-up experiment, we explicitly instruct the model to prioritize careful execution of its role and to focus on being the best-performing informant in the given task.\footnote{\texttt{"role": "system", "content": "Tu participes à une étude linguistique et tu es de langue maternelle française. Ton objectif premier est de te concentrer sur ta tâche, et de répondre attentivement."}, in English: You are a participant in a language study and your mother tongue is French. Your primary objective is to concentrate on your task, and to respond with attention to detail.} The accuracy scores across the four conditions tested and on filler sentences are shown in Appendix~\ref{appendix:B}. The model keeps outperforming human participants and reaches high accuracy rates (M-GPT-4o-mini-role= 0.98). Condition 4 registers the lowest score (M-GPT-4o-mini-role= 0.94), which is in line with our baseline. The performance of GPT-4o-mini on filler sentences improves of 0.06 points.

\subsubsection{Follow-up 2: Chain-of-Thought (CoT) prompting}
Second, we introduce a different modification to the original pipeline, independent of the first follow-up experiment, and we implement Chain-of-Thought (CoT) prompting. The modification of the prompt aims to encourage the model to reason through its decisions more explicitly. CoT elicits a series of intermediate reasoning steps from the model to improve the ability of LLMs to perform complex reasoning \citep{wei_chain--thought_2022}. Two examples, one from the critical and one from the filler stimuli, are shown in the adapted prompt to exemplify how the thinking process should take place in order to solve the task at best. Both example sentences are excluded from the test dataset. A summary of the versions of used prompts across the different study replication designs can be found in Appendix~\ref{appendix:C}. 

CoT results present a general small drop in accuracy (Conditions 2, 3, and 4) in respect to our baseline and the first follow-up experiment with the modified zero-shot prompting function. However, it maintains high accuracy scores (M-GPT-4o-mini-CoT= 0.94). In Condition 4, human participants outperform the model for 0,03 points (M-human= 0.92, M-GPT-4o-mini-CoT= 0.91), but the two scores are close. The gain of this second follow-up experiments is GPT-4o-mini's performance on filler sentences, reaching 0.99 accuracy. 

\section{Discussion}
\label{sec:discu}

This study examined the performance of LLMs in replicating empirical linguistic studies originally conducted with human participants. While our analysis focuses on two specific cases, we demonstrate that this approach is suitable across a broad range of disciplines both within and beyond the humanities. Our findings highlight both the potential and the limitations of using LLMs as the future crowd workers of empirical Linguistics. The major takeaway from this paper is a general tendency to alignments between human and LLM judgments in linguistic tasks, suggesting that these models serve as a solid starting point for further research on synthetic crowdsourcing studies. This can be carried out also by non-experts as our accessible code base exhibits and can be extended to a wide range of applications, as demonstrated in our follow-up experiments.

Breaking down the findings from the two replications, the main constant is a very accurate task completion carried out by  GPT-4o-mini. The model outperforms human participants in both case studies across all conditions tested. The reproduction of Cruz\_23 supports the notion that the gender of the Spanish translation equivalent (analogical gender) of the English target noun mediates the assignment mechanism in code-switched speech also for LLMs. The tendencies observed in GPT-4o-mini's performance are numerically close to those registered with human participants. First, human-denoting nouns appear to evoke the semantic (biological) gender of their referents, reaching the highest gender congruency rates in the replication. While for humans this effect is linked to real-world referential properties, LLMs lack direct access to the real world, implying that such properties must be encoded in their vast training corpus. Second, Spanish-English bilinguals typically default to a masculine assignment strategy for inanimate nouns, a trend that is replicated in our study. Third, phonemic cues in the morphological endings of Spanish translations of English target nouns play a facilitating role, especially for LLMs, in assigning gender, particularly in the case of feminine inanimate nouns.

In contrast, the second replication (Lombard\_21) presents a different scenario in three key aspects. First, after removing outliers, no significant differences between experimental conditions are observed, leading to the rejection of the original study’s hypothesis that neologism detection may be influenced by neologism formation and linguistic usage patterns. Second, the measurement of reaction times in GPT-4o-mini raises concerns regarding using this parameter in future replications since it may be influenced by external factors that may affect computation. Especially its consistency as a variable of meaningful comparability across iterations is questioned, highlighting the need for further research into alternative evaluation methods for human vs. LLMs performance. Third, while the model excels in detecting neologisms, its accuracy drops significantly in identifying fillers. This finding proposes two considerations: first, zero-shot prompting may not be the optimal strategy for all experimental scenarios \citep{van_dalfsen_direct_2024}, reinforcing the importance of exploring alternative prompting techniques. The first follow-up experiment reveals that the role parameter in OpenAI chat models' zero-shot prompting function does not significantly impact task outcomes. The second experiment demonstrates that Chain-of-Thought (CoT) prompting is a viable alternative to zero-shot prompting for more structured tasks, as it enhances complex reasoning \citep{wei_chain--thought_2022}. The model’s performance in the CoT experiment aligns more closely with that of human participants, whose results are high but not as consistently high as the model's baseline performance.

This generally high performance of LLMs can be interpreted in two ways. On the one hand, it serves as an encouragement to integrate these models into research pipelines beyond the NLP community. On the other hand, it raises concerns about potential "over-performance" of those models. LLMs are trained to be highly efficient assistants, often avoiding expressions of uncertainty or negative responses. This may explain also why "yes" answers may be preferred to "no" answers since they may convey a better user-experience. The second replication and its follow-up experiments highlight that LLMs do not consistently serve as suitable substitutes for human participants, as the differences observed in human ratings are often neutralized or even reversed. While GPT-4o-mini exhibits superior performance in detecting neologisms, this advantage becomes a limitation in the context of the original study, which seeks to examine human strategies for identifying new words. The model’s high accuracy, particularly in Conditions (2) and (3), where human participants demonstrated lower detection rates, indicates that it may rely on different underlying mechanisms than those employed by humans. However, when the task is structured as a sequence of reasoning steps (CoT), as implemented in our second follow-up experiment, the model's responses improve in accuracy on filler items. This underscores the potential necessity of a greater adaptation of linguistic research methodologies in their replication with LLMs, as approaches effective for human participants may not be directly transferable to LLM-based pipelines.

\section{Conclusion}
\label{sec:conclu}
This work introduces a framework to replicate empirical linguistic tasks, which are carried out with human participants, with the help of LLMs. Under this framework, OpenAI's GPT-4o-mini is implied to reproduce two forced tasks with different linguistic foci. Two general tendencies arise in the analysis of reproductions outcomes. First, the empirical results demonstrate the effectiveness of the proposed framework in achieving great task accuracies. GPT-4o-mini outperforms human participants in all experimental conditions tested. Second, alignment patterns in LLMs' decision-making processes with human judgments are outlined especially in the first replication. The second task reproduced shows that GPT-4o-mini may overperfom and not be always an adequate substitute of human participants. However, the two follow-up experiments demonstrate the adaptation and extension potential of our research design, focusing on ways to refine model's outputs.
Overall, our framework has some limitations. The number of replications, linguistic subfields, and model's architectures tested in the present paper is restricted and should constitute a starting point for further research. Future studies could also test the replication design with open-source LLMs to have better control over the availability of data for the model. In conclusion, our replications use only one model, GPT-4o-mini, which, despite being one of the newest and most powerful of the OpenAI family, does not provide the opportunity for further comparisons. 

\bibliographystyle{plainnat}
\bibliography{CHR_paper_1}  

\begin{thebibliography}{36}
\providecommand{\natexlab}[1]{#1}
\providecommand{\url}[1]{\texttt{#1}}
\expandafter\ifx\csname urlstyle\endcsname\relax
  \providecommand{\doi}[1]{doi: #1}\else
  \providecommand{\doi}{doi: \begingroup \urlstyle{rm}\Url}\fi

\bibitem[Ahmadabadi et~al.()Ahmadabadi, Haghifam, Shah-Mansouri, and Ershadmanesh]{ahmadabadi_2024}
Samin~Nili Ahmadabadi, Maryam Haghifam, Vahid Shah-Mansouri, and Sara Ershadmanesh.
\newblock Design and evaluation of crowdsourcing platforms based on users’ confidence judgments.
\newblock 14\penalty0 (1):\penalty0 18379.
\newblock ISSN 2045-2322.
\newblock \doi{10.1038/s41598-024-65892-7}.
\newblock URL \url{https://www.nature.com/articles/s41598-024-65892-7}.

\bibitem[Alizadeh et~al.(2025)Alizadeh, Kubli, Samei, Dehghani, Zahedivafa, Bermeo, Korobeynikova, and Gilardi]{alizadeh2025-cm}
Meysam Alizadeh, Ma{\"e}l Kubli, Zeynab Samei, Shirin Dehghani, Mohammadmasiha Zahedivafa, Juan~D Bermeo, Maria Korobeynikova, and Fabrizio Gilardi.
\newblock Open-source {LLMs} for text annotation: a practical guide for model setting and fine-tuning.
\newblock \emph{J. Comput. Soc. Sci.}, 8\penalty0 (1):\penalty0 17, 2025.

\bibitem[Argyle et~al.(2023)Argyle, Busby, Fulda, Gubler, Rytting, and Wingate]{Argyle_2023}
Lisa~P. Argyle, Ethan~C. Busby, Nancy Fulda, Joshua~R. Gubler, Christopher Rytting, and David Wingate.
\newblock Out of one, many: Using language models to simulate human samples.
\newblock \emph{Political Analysis}, 31\penalty0 (3):\penalty0 337–351, February 2023.
\newblock ISSN 1476-4989.
\newblock \doi{10.1017/pan.2023.2}.
\newblock URL \url{http://dx.doi.org/10.1017/pan.2023.2}.

\bibitem[Borst et~al.()Borst, Klähn, and Burghardt]{borst_death_2023}
Janos Borst, Jannis Klähn, and Manuel Burghardt.
\newblock Death of the dictionary? – the rise of zero-shot sentiment classification.
\newblock volume {CEUR} Workshop Proceedings.
\newblock URL \url{https://ceur-ws.org/Vol-3558/paper3130.pdf}.

\bibitem[Capuano et~al.()Capuano, Dudschig, Günther, and Kaup]{capuano_semantic_2021}
Francesca Capuano, Carolin Dudschig, Fritz Günther, and Barbara Kaup.
\newblock Semantic similarity of alternatives fostered by conversational negation.
\newblock 45\penalty0 (7):\penalty0 e13015.
\newblock ISSN 0364-0213, 1551-6709.
\newblock \doi{10.1111/cogs.13015}.
\newblock URL \url{https://onlinelibrary.wiley.com/doi/10.1111/cogs.13015}.

\bibitem[Carvalho()]{carvalho_corpus_2012}
A~Carvalho.
\newblock Corpus del español en el sur de arizona ({CESA}).

\bibitem[Cegin et~al.()Cegin, Simko, and Brusilovsky]{cegin_chatgpt_2023}
Jan Cegin, Jakub Simko, and Peter Brusilovsky.
\newblock {ChatGPT} to replace crowdsourcing of paraphrases for intent classification: Higher diversity and comparable model robustness.
\newblock In \emph{Proceedings of the 2023 Conference on Empirical Methods in Natural Language Processing}, pages 1889--1905. Association for Computational Linguistics.
\newblock \doi{10.18653/v1/2023.emnlp-main.117}.
\newblock URL \url{https://aclanthology.org/2023.emnlp-main.117}.

\bibitem[Cruz()]{cruz2023}
Abel Cruz.
\newblock Linguistic factors modulating gender assignment in spanish–english bilingual speech.
\newblock 26\penalty0 (3):\penalty0 580--591.
\newblock ISSN 1366-7289, 1469-1841.
\newblock \doi{10.1017/S1366728922000839}.
\newblock URL \url{https://www.cambridge.org/core/product/identifier/S1366728922000839/type/journal_article}.

\bibitem[De~Lange et~al.()De~Lange, Vanroy, De~Bruyne, Singh, Lefever, and De~Clercq]{de_lange_benchmarking_2024}
Loic De~Lange, Bram Vanroy, Luna De~Bruyne, Pranaydeep Singh, Els Lefever, and Orphée De~Clercq.
\newblock Benchmarking zero-shot text classification for dutch.
\newblock Computational Linguistics in the Netherlands Journal\penalty0 (13):\penalty0 63--90.
\newblock URL \url{https://clinjournal.org/clinj/article /view/172}.

\bibitem[Dillion et~al.()Dillion, Tandon, Gu, and Gray]{dillion_can_2023}
Danica Dillion, Niket Tandon, Yuling Gu, and Kurt Gray.
\newblock Can {AI} language models replace human participants?
\newblock 27\penalty0 (7):\penalty0 597--600.
\newblock ISSN 13646613.
\newblock \doi{10.1016/j.tics.2023.04.008}.
\newblock URL \url{https://linkinghub.elsevier.com/retrieve/pii/S1364661323000980}.

\bibitem[Duncan()]{duncan_does_2024}
Daniel Duncan.
\newblock Does {ChatGPT} have sociolinguistic competence?
\newblock 8:\penalty0 51--75.
\newblock ISSN 2530-9455.
\newblock \doi{10.4995/jclr.2024.21958}.
\newblock URL \url{https://polipapers.upv.es/index.php/jclr/article/view/21958}.

\bibitem[Dynel()]{dynel_lessons_2023}
Marta Dynel.
\newblock Lessons in linguistics with {ChatGPT}: Metapragmatics, metacommunication, metadiscourse and metalanguage in human-{AI} interactions.
\newblock 93:\penalty0 107--124.
\newblock ISSN 02715309.
\newblock \doi{10.1016/j.langcom.2023.09.002}.
\newblock URL \url{https://linkinghub.elsevier.com/retrieve/pii/S0271530923000605}.

\bibitem[Gilardi et~al.()Gilardi, Alizadeh, and Kubli]{gilardi_chatgpt_2023}
Fabrizio Gilardi, Meysam Alizadeh, and Maël Kubli.
\newblock {ChatGPT} outperforms crowd-workers for text-annotation tasks.
\newblock \doi{10.48550/ARXIV.2303.15056}.
\newblock URL \url{https://arxiv.org/abs/2303.15056}.
\newblock Publisher: {arXiv} Version Number: 2.

\bibitem[Gui and Toubia()]{gui_challenge_2023}
George Gui and Olivier Toubia.
\newblock The challenge of using {LLMs} to simulate human behavior: A causal inference perspective.
\newblock \doi{10.48550/ARXIV.2312.15524}.
\newblock URL \url{https://arxiv.org/abs/2312.15524}.
\newblock Publisher: {arXiv} Version Number: 2.

\bibitem[Han()]{han_chatgpt_2024}
{ZhaoHong} Han.
\newblock Chatgpt in and for second language acquisition: A call for systematic research.
\newblock 46\penalty0 (2):\penalty0 301--306.
\newblock ISSN 0272-2631, 1470-1545.
\newblock \doi{10.1017/S0272263124000111}.
\newblock URL \url{https://www.cambridge.org/core/product/identifier/S0272263124000111/type/journal_article}.

\bibitem[He et~al.()He, Lin, Gong, Jin, Zhang, Lin, Jiao, Yiu, Duan, and Chen]{he_annollm_2023}
Xingwei He, Zhenghao Lin, Yeyun Gong, A-Long Jin, Hang Zhang, Chen Lin, Jian Jiao, Siu~Ming Yiu, Nan Duan, and Weizhu Chen.
\newblock {AnnoLLM}: Making large language models to be better crowdsourced annotators.
\newblock URL \url{https://arxiv.org/abs/2303.16854}.
\newblock Version Number: 2.

\bibitem[Horych et~al.()Horych, Mandl, Ruas, Greiner-Petter, Gipp, Aizawa, and Spinde]{horych_promises_2024}
Tomas Horych, Christoph Mandl, Terry Ruas, Andre Greiner-Petter, Bela Gipp, Akiko Aizawa, and Timo Spinde.
\newblock The promises and pitfalls of {LLM} annotations in dataset labeling: a case study on media bias detection.
\newblock URL \url{https://arxiv.org/abs/2411.11081}.
\newblock Version Number: 1.

\bibitem[Karjus(2024)]{karjus2024machineassisted}
Andres Karjus.
\newblock Machine-assisted quantitizing designs: augmenting humanities and social sciences with artificial intelligence, 2024.
\newblock URL \url{https://arxiv.org/abs/2309.14379}.

\bibitem[Karjus and Cuskley(2024)]{Karjus2024-zg}
Andres Karjus and Christine Cuskley.
\newblock Evolving linguistic divergence on polarizing social media.
\newblock \emph{Humanit. Soc. Sci. Commun.}, 11\penalty0 (1), March 2024.

\bibitem[Khalilia et~al.()Khalilia, Otterbacher, Bella, Noortyani, Darma, and Giunchiglia]{khalilia_crowdsourcing_2024}
Hadi Khalilia, Jahna Otterbacher, Gabor Bella, Rusma Noortyani, Shandy Darma, and Fausto Giunchiglia.
\newblock Crowdsourcing lexical diversity.
\newblock URL \url{https://arxiv.org/abs/2410.23133}.
\newblock Version Number: 1.

\bibitem[Kocoń et~al.()Kocoń, Cichecki, Kaszyca, Kochanek, Szydło, Baran, Bielaniewicz, Gruza, Janz, Kanclerz, Kocoń, Koptyra, Mieleszczenko-Kowszewicz, Miłkowski, Oleksy, Piasecki, Radliński, Wojtasik, Woźniak, and Kazienko]{kocon_chatgpt_2023}
Jan Kocoń, Igor Cichecki, Oliwier Kaszyca, Mateusz Kochanek, Dominika Szydło, Joanna Baran, Julita Bielaniewicz, Marcin Gruza, Arkadiusz Janz, Kamil Kanclerz, Anna Kocoń, Bartłomiej Koptyra, Wiktoria Mieleszczenko-Kowszewicz, Piotr Miłkowski, Marcin Oleksy, Maciej Piasecki, Łukasz Radliński, Konrad Wojtasik, Stanisław Woźniak, and Przemysław Kazienko.
\newblock {ChatGPT}: Jack of all trades, master of none.
\newblock 99:\penalty0 101861.
\newblock ISSN 15662535.
\newblock \doi{10.1016/j.inffus.2023.101861}.
\newblock URL \url{https://linkinghub.elsevier.com/retrieve/pii/S156625352300177X}.

\bibitem[Kosar et~al.()Kosar, De~Pauw, and Daelemans]{kosar_comparative_2024}
A~Kosar, G~De~Pauw, and W~Daelemans.
\newblock Comparative evaluation of topic detection: Humans vs. {LLMs}.
\newblock Computational Linguistics in the Netherlands Journal\penalty0 (13):\penalty0 91--120.
\newblock URL \url{https://www.clinjournal.org/clinj/article/view/173}.

\bibitem[Lombard et~al.()Lombard, Huyghe, and Gygax]{lombard2021}
Alizée Lombard, Richard Huyghe, and Pascal Gygax.
\newblock Neological intuition in french: A study of formal novelty and lexical regularity as predictors.
\newblock 254:\penalty0 103055.
\newblock ISSN 00243841.
\newblock \doi{10.1016/j.lingua.2021.103055}.
\newblock URL \url{https://linkinghub.elsevier.com/retrieve/pii/S0024384121000279}.

\bibitem[OpenAI et~al.()OpenAI, Achiam, Adler, Agarwal, Ahmad, Akkaya, Aleman, Almeida, Altenschmidt, Altman, Anadkat, Avila, Babuschkin, Balaji, Balcom, Baltescu, Bao, Bavarian, Belgum, Bello, Berdine, Bernadett-Shapiro, Berner, Bogdonoff, Boiko, Boyd, Brakman, Brockman, Brooks, Brundage, Button, Cai, Campbell, Cann, Carey, Carlson, Carmichael, Chan, Chang, Chantzis, Chen, Chen, Chen, Chen, Chen, Chess, Cho, Chu, Chung, Cummings, Currier, Dai, Decareaux, Degry, Deutsch, Deville, Dhar, Dohan, Dowling, Dunning, Ecoffet, Eleti, Eloundou, Farhi, Fedus, Felix, Fishman, Forte, Fulford, Gao, Georges, Gibson, Goel, Gogineni, Goh, Gontijo-Lopes, Gordon, Grafstein, Gray, Greene, Gross, Gu, Guo, Hallacy, Han, Harris, He, Heaton, Heidecke, Hesse, Hickey, Hickey, Hoeschele, Houghton, Hsu, Hu, Hu, Huizinga, Jain, Jain, Jang, Jiang, Jiang, Jin, Jin, Jomoto, Jonn, Jun, Kaftan, Kaiser, Kamali, Kanitscheider, Keskar, Khan, Kilpatrick, Kim, Kim, Kim, Kirchner, Kiros, Knight, Kokotajlo, Kondraciuk, Kondrich, Konstantinidis,
  Kosic, Krueger, Kuo, Lampe, Lan, Lee, Leike, Leung, Levy, Li, Lim, Lin, Lin, Litwin, Lopez, Lowe, Lue, Makanju, Malfacini, Manning, Markov, Markovski, Martin, Mayer, Mayne, McGrew, McKinney, McLeavey, McMillan, McNeil, Medina, Mehta, Menick, Metz, Mishchenko, Mishkin, Monaco, Morikawa, Mossing, Mu, Murati, Murk, Mély, Nair, Nakano, Nayak, Neelakantan, Ngo, Noh, Ouyang, O'Keefe, Pachocki, Paino, Palermo, Pantuliano, Parascandolo, Parish, Parparita, Passos, Pavlov, Peng, Perelman, Peres, Petrov, Pinto, Michael, {Pokorny}, Pokrass, Pong, Powell, Power, Power, Proehl, Puri, Radford, Rae, Ramesh, Raymond, Real, Rimbach, Ross, Rotsted, Roussez, Ryder, Saltarelli, Sanders, Santurkar, Sastry, Schmidt, Schnurr, Schulman, Selsam, Sheppard, Sherbakov, Shieh, Shoker, Shyam, Sidor, Sigler, Simens, Sitkin, Slama, Sohl, Sokolowsky, Song, Staudacher, Such, Summers, Sutskever, Tang, Tezak, Thompson, Tillet, Tootoonchian, Tseng, Tuggle, Turley, Tworek, Uribe, Vallone, Vijayvergiya, Voss, Wainwright, Wang, Wang, Wang, Ward,
  Wei, Weinmann, Welihinda, Welinder, Weng, Weng, Wiethoff, Willner, Winter, Wolrich, Wong, Workman, Wu, Wu, Wu, Xiao, Xu, Yoo, Yu, Yuan, Zaremba, Zellers, Zhang, Zhang, Zhao, Zheng, Zhuang, Zhuk, and Zoph]{openai_gpt-4_2023}
OpenAI, Josh Achiam, Steven Adler, Sandhini Agarwal, Lama Ahmad, Ilge Akkaya, Florencia~Leoni Aleman, Diogo Almeida, Janko Altenschmidt, Sam Altman, Shyamal Anadkat, Red Avila, Igor Babuschkin, Suchir Balaji, Valerie Balcom, Paul Baltescu, Haiming Bao, Mohammad Bavarian, Jeff Belgum, Irwan Bello, Jake Berdine, Gabriel Bernadett-Shapiro, Christopher Berner, Lenny Bogdonoff, Oleg Boiko, Madelaine Boyd, Anna-Luisa Brakman, Greg Brockman, Tim Brooks, Miles Brundage, Kevin Button, Trevor Cai, Rosie Campbell, Andrew Cann, Brittany Carey, Chelsea Carlson, Rory Carmichael, Brooke Chan, Che Chang, Fotis Chantzis, Derek Chen, Sully Chen, Ruby Chen, Jason Chen, Mark Chen, Ben Chess, Chester Cho, Casey Chu, Hyung~Won Chung, Dave Cummings, Jeremiah Currier, Yunxing Dai, Cory Decareaux, Thomas Degry, Noah Deutsch, Damien Deville, Arka Dhar, David Dohan, Steve Dowling, Sheila Dunning, Adrien Ecoffet, Atty Eleti, Tyna Eloundou, David Farhi, Liam Fedus, Niko Felix, Simón~Posada Fishman, Juston Forte, Isabella Fulford, Leo
  Gao, Elie Georges, Christian Gibson, Vik Goel, Tarun Gogineni, Gabriel Goh, Rapha Gontijo-Lopes, Jonathan Gordon, Morgan Grafstein, Scott Gray, Ryan Greene, Joshua Gross, Shixiang~Shane Gu, Yufei Guo, Chris Hallacy, Jesse Han, Jeff Harris, Yuchen He, Mike Heaton, Johannes Heidecke, Chris Hesse, Alan Hickey, Wade Hickey, Peter Hoeschele, Brandon Houghton, Kenny Hsu, Shengli Hu, Xin Hu, Joost Huizinga, Shantanu Jain, Shawn Jain, Joanne Jang, Angela Jiang, Roger Jiang, Haozhun Jin, Denny Jin, Shino Jomoto, Billie Jonn, Heewoo Jun, Tomer Kaftan, Łukasz Kaiser, Ali Kamali, Ingmar Kanitscheider, Nitish~Shirish Keskar, Tabarak Khan, Logan Kilpatrick, Jong~Wook Kim, Christina Kim, Yongjik Kim, Jan~Hendrik Kirchner, Jamie Kiros, Matt Knight, Daniel Kokotajlo, Łukasz Kondraciuk, Andrew Kondrich, Aris Konstantinidis, Kyle Kosic, Gretchen Krueger, Vishal Kuo, Michael Lampe, Ikai Lan, Teddy Lee, Jan Leike, Jade Leung, Daniel Levy, Chak~Ming Li, Rachel Lim, Molly Lin, Stephanie Lin, Mateusz Litwin, Theresa Lopez, Ryan
  Lowe, Patricia Lue, Anna Makanju, Kim Malfacini, Sam Manning, Todor Markov, Yaniv Markovski, Bianca Martin, Katie Mayer, Andrew Mayne, Bob McGrew, Scott~Mayer McKinney, Christine McLeavey, Paul McMillan, Jake McNeil, David Medina, Aalok Mehta, Jacob Menick, Luke Metz, Andrey Mishchenko, Pamela Mishkin, Vinnie Monaco, Evan Morikawa, Daniel Mossing, Tong Mu, Mira Murati, Oleg Murk, David Mély, Ashvin Nair, Reiichiro Nakano, Rajeev Nayak, Arvind Neelakantan, Richard Ngo, Hyeonwoo Noh, Long Ouyang, Cullen O'Keefe, Jakub Pachocki, Alex Paino, Joe Palermo, Ashley Pantuliano, Giambattista Parascandolo, Joel Parish, Emy Parparita, Alex Passos, Mikhail Pavlov, Andrew Peng, Adam Perelman, Filipe de Avila~Belbute Peres, Michael Petrov, Henrique Ponde de~Oliveira Pinto, Michael, {Pokorny}, Michelle Pokrass, Vitchyr~H. Pong, Tolly Powell, Alethea Power, Boris Power, Elizabeth Proehl, Raul Puri, Alec Radford, Jack Rae, Aditya Ramesh, Cameron Raymond, Francis Real, Kendra Rimbach, Carl Ross, Bob Rotsted, Henri Roussez,
  Nick Ryder, Mario Saltarelli, Ted Sanders, Shibani Santurkar, Girish Sastry, Heather Schmidt, David Schnurr, John Schulman, Daniel Selsam, Kyla Sheppard, Toki Sherbakov, Jessica Shieh, Sarah Shoker, Pranav Shyam, Szymon Sidor, Eric Sigler, Maddie Simens, Jordan Sitkin, Katarina Slama, Ian Sohl, Benjamin Sokolowsky, Yang Song, Natalie Staudacher, Felipe~Petroski Such, Natalie Summers, Ilya Sutskever, Jie Tang, Nikolas Tezak, Madeleine~B. Thompson, Phil Tillet, Amin Tootoonchian, Elizabeth Tseng, Preston Tuggle, Nick Turley, Jerry Tworek, Juan Felipe~Cerón Uribe, Andrea Vallone, Arun Vijayvergiya, Chelsea Voss, Carroll Wainwright, Justin~Jay Wang, Alvin Wang, Ben Wang, Jonathan Ward, Jason Wei, CJ~Weinmann, Akila Welihinda, Peter Welinder, Jiayi Weng, Lilian Weng, Matt Wiethoff, Dave Willner, Clemens Winter, Samuel Wolrich, Hannah Wong, Lauren Workman, Sherwin Wu, Jeff Wu, Michael Wu, Kai Xiao, Tao Xu, Sarah Yoo, Kevin Yu, Qiming Yuan, Wojciech Zaremba, Rowan Zellers, Chong Zhang, Marvin Zhang, Shengjia
  Zhao, Tianhao Zheng, Juntang Zhuang, William Zhuk, and Barret Zoph.
\newblock {GPT}-4 technical report.
\newblock URL \url{https://arxiv.org/abs/2303.08774}.
\newblock Version Number: 6.

\bibitem[Ostyakova et~al.()Ostyakova, Smilga, Petukhova, Molchanova, and Kornev]{ostyakova_chatgpt_2023}
Lidiia Ostyakova, Veronika Smilga, Kseniia Petukhova, Maria Molchanova, and Daniel Kornev.
\newblock {ChatGPT} vs. crowdsourcing vs. experts: Annotating open-domain conversations with speech functions.
\newblock In \emph{Proceedings of the 24th Meeting of the Special Interest Group on Discourse and Dialogue}, pages 242--254. Association for Computational Linguistics.
\newblock \doi{10.18653/v1/2023.sigdial-1.23}.
\newblock URL \url{https://aclanthology.org/2023.sigdial-1.23}.

\bibitem[Rebora et~al.()Rebora, Lehmann, Heumann, Ding, and Lauer]{rebora_comparing_2023}
Simone Rebora, Marina Lehmann, Anne Heumann, Wei Ding, and Gerhard Lauer.
\newblock Comparing {ChatGPT} to human raters and sentiment analysis tools for german children’s literature.
\newblock volume {CEUR} Workshop Proceedings.
\newblock URL \url{https://ceur-ws.org/Vol-3558/paper3340.pdf}.

\bibitem[Sheehan et~al.()Sheehan, Schäfer, and Parafita~Couto]{sheehan_crowdsourcing_2019}
Michelle Sheehan, Martin Schäfer, and Maria~Carmen Parafita~Couto.
\newblock Crowdsourcing and minority languages: The case of galician inflected infinitives1.
\newblock 10:\penalty0 1157.
\newblock ISSN 1664-1078.
\newblock \doi{10.3389/fpsyg.2019.01157}.
\newblock URL \url{https://www.frontiersin.org/article/10.3389/fpsyg.2019.01157/full}.

\bibitem[Thwaites et~al.()Thwaites, Vandeweerd, and Paquot]{thwaites_crowdsourced_2024}
Peter Thwaites, Nathan Vandeweerd, and Magali Paquot.
\newblock Crowdsourced comparative judgement for evaluating learner texts: How reliable are judges recruited from an online crowdsourcing platform?
\newblock page amae048.
\newblock ISSN 0142-6001, 1477-450X.
\newblock \doi{10.1093/applin/amae048}.
\newblock URL \url{https://academic.oup.com/applij/advance-article/doi/10.1093/applin/amae048/7719043}.

\bibitem[Törnberg()]{tornberg_chatgpt-4_2023}
Petter Törnberg.
\newblock {ChatGPT}-4 outperforms experts and crowd workers in annotating political twitter messages with zero-shot learning.
\newblock URL \url{https://arxiv.org/abs/2304.06588}.
\newblock Version Number: 1.

\bibitem[Van~Dalfsen et~al.()Van~Dalfsen, Karsdorp, Bagheri, Mentink, van Engelen, and Stronks]{van_dalfsen_direct_2024}
Arjan Van~Dalfsen, Folgert Karsdorp, Ayoub Bagheri, Dieuwertje Mentink, Thirza van Engelen, and Els Stronks.
\newblock Direct and indirect annotation with generative {AI}: A case study into finding animals and plants in historical text.
\newblock volume {CEUR} Workshop Proceedings.
\newblock URL \url{https://ceur-ws.org/Vol-3834/paper74.pdf}.

\bibitem[Van~Zoonen et~al.()Van~Zoonen, Sivunen, and Treem]{van_zoonen_algorithmic_2024}
Ward Van~Zoonen, Anu~E. Sivunen, and Jeffrey~W. Treem.
\newblock Algorithmic management of crowdworkers: Implications for workers’ identity, belonging, and meaningfulness of work.
\newblock 152:\penalty0 108089.
\newblock ISSN 07475632.
\newblock \doi{10.1016/j.chb.2023.108089}.
\newblock URL \url{https://linkinghub.elsevier.com/retrieve/pii/S0747563223004405}.

\bibitem[Veselovsky et~al.()Veselovsky, Ribeiro, and West]{veselovsky_artificial_2023}
Veniamin Veselovsky, Manoel~Horta Ribeiro, and Robert West.
\newblock Artificial artificial artificial intelligence: Crowd workers widely use large language models for text production tasks.
\newblock URL \url{https://arxiv.org/abs/2306.07899}.
\newblock Version Number: 1.

\bibitem[Wei et~al.()Wei, Wang, Schuurmans, Bosma, Ichter, Xia, Chi, Le, and Zhou]{wei_chain--thought_2022}
Jason Wei, Xuezhi Wang, Dale Schuurmans, Maarten Bosma, Brian Ichter, Fei Xia, Ed~Chi, Quoc Le, and Denny Zhou.
\newblock Chain-of-thought prompting elicits reasoning in large language models.
\newblock URL \url{https://arxiv.org/abs/2201.11903}.
\newblock Version Number: 6.

\bibitem[Wu et~al.()Wu, Zhu, Albayrak, Axon, Bertsch, Deng, Ding, Guo, Gururaja, Kuo, Liang, Liu, Mandal, Milbauer, Ni, Padmanabhan, Ramkumar, Sudjianto, Taylor, Tseng, Vaidos, Wu, Wu, and Yang]{wu_llms_2025}
Tongshuang Wu, Haiyi Zhu, Maya Albayrak, Alexis Axon, Amanda Bertsch, Wenxing Deng, Ziqi Ding, Bill Guo, Sireesh Gururaja, Tzu-Sheng Kuo, Jenny~T. Liang, Ryan Liu, Ihita Mandal, Jeremiah Milbauer, Xiaolin Ni, Namrata Padmanabhan, Subhashini Ramkumar, Alexis Sudjianto, Jordan Taylor, Ying-Jui Tseng, Patricia Vaidos, Zhijin Wu, Wei Wu, and Chenyang Yang.
\newblock {LLMs} as workers in human-computational algorithms? replicating crowdsourcing pipelines with {LLMs}.
\newblock ISBN 79-8-4007-1395-8/25/04.
\newblock \doi{10.48550/ARXIV.2307.10168}.
\newblock URL \url{https://arxiv.org/abs/2307.10168}.
\newblock Publisher: {arXiv} Version Number: 3.

\bibitem[Xu et~al.()Xu, Han, Sadiq, and Demartini]{xu_role_2024}
Jiechen Xu, Lei Han, Shazia Sadiq, and Gianluca Demartini.
\newblock On the role of large language models in crowdsourcing misinformation assessment.
\newblock 18:\penalty0 1674--1686.
\newblock ISSN 2334-0770, 2162-3449.
\newblock \doi{10.1609/icwsm.v18i1.31417}.
\newblock URL \url{https://ojs.aaai.org/index.php/ICWSM/article/view/31417}.

\bibitem[Ziems et~al.()Ziems, Held, Shaikh, Chen, Zhang, and Yang]{ziems_can_2024}
Caleb Ziems, William Held, Omar Shaikh, Jiaao Chen, Zhehao Zhang, and Diyi Yang.
\newblock Can large language models transform computational social science?
\newblock 50\penalty0 (1):\penalty0 237--291.
\newblock ISSN 0891-2017, 1530-9312.
\newblock \doi{10.1162/coli_a_00502}.
\newblock URL \url{https://direct.mit.edu/coli/article/50/1/237/118498/Can-Large-Language-Models-Transform-Computational}.

\end{thebibliography}

\newpage
\appendix
  
\section{Appendix A: Supplementary materials}
\label{appendix:A}
The Python notebook that contains the code used for in present paper can be accessed at this \href{https://github.com/irisferrazzo/LLMs_as_crowdworkers}{link}. The supplementary materials of \citet{cruz2023} and \citet{lombard2021} were reorganized in a new format that can be found \href{https://anonymous.4open.science/r/LLMs_as_crowdworkers/README.md}{here}.

\section{Appendix B: Results from the conducted replications.}
\label{appendix:B}

\begin{table}[hbt!]
\centering
\caption{Results from the cross-validation of replication 1: Cruz\_23}
\resizebox{\columnwidth}{!}{
\begin{tabular}{l p{0.15\textwidth} p{0.5\textwidth} p{0.15\textwidth}} 
\toprule
\textbf{Condition} & \textbf{Accuracy} & \textbf{Wrong targets} & \textbf{Num. errors} \\ 
\midrule
condition\_1\_1 & 0.98 & \textit{hijo} (son), \textit{abuelo} (grandpa), \textit{bombero} (fireman) & 5  \\ 

condition\_2\_1 & 0.99 & \textit{hombre} (man), \textit{ayudante} (busboy) & 2 \\ 

condition\_3\_1 & 0.99 & \textit{techo} (roof), \textit{pueblo} (town) & 3 \\ 

condition\_4\_1 & 0.93 & \textit{tenedor} (fork), \textit{reloj} (clock) & 22 \\

condition\_5\_1 & 1 & / & / \\ 

condition\_6\_1 & 1 & / & / \\ 

condition\_7\_1 & 0.88 & \textit{ventana} (window), \textit{receta} (recipe), \textit{mesa} (table), \textit{frontera} (border), \textit{valla} (fence) & 41 \\

condition\_8\_1 & 0.67 & \textit{ley} (law), \textit{cárcel} (jail), \textit{miel} (honey), \textit{leche} (milk), \textit{sal} (salt), \textit{nieve} (snow), \textit{tos} (cough), \textit{cruz} (cross) & 111 \\

condition\_1\_2 & 0.97 & \textit{hijo} (son), \textit{bombero} (fireman) & 9  \\ 

condition\_2\_2 & 0.99 & \textit{hombre} (man), \textit{rey} (king) & 3 \\ 

condition\_3\_2 & 0.99 & \textit{techo} (roof) & 1 \\ 

condition\_4\_2 & 0.96 & \textit{tenedor} (fork), \textit{reloj} (clock) & 12 \\

condition\_5\_2 & 0.99 & \textit{monja} (nun) & 1 \\ 

condition\_6\_2 & 1 & / & / \\ 

condition\_7\_2 & 0.89 & \textit{mesa} (table), \textit{frontera} (border), \textit{valla} (fence) & 37 \\

condition\_8\_2 & 0.79 & \textit{ciudad} (town), \textit{cárcel} (jail), \textit{beca} (grant), \textit{leche} (milk) & 69 \\

\bottomrule
\end{tabular}%
}
\end{table}

\begin{table}[hbt!]
\centering
\caption{Accuracy results from all experiments within replication 2: Lombard\_21}
\fontsize{11}{13}\selectfont 
\resizebox{\columnwidth}{!}{
\begin{tabular}{lp{0.2\textwidth}p{0.2\textwidth}p{0.2\textwidth}p{0.2\textwidth}p{0.2\textwidth}} 
\toprule
\textbf{Condition} & \textbf{zero-shot-1} & \textbf{zero-shot-2} & \textbf{zero-shot-role} & \textbf{CoT}\\ 
\midrule
condition\_1 & 0.99 & 0.98 & 1 & 0.99  \\ 

condition\_2 & 0.99 & 0.99 & 0.99 & 0.91 \\ 

condition\_3 & 0.99 & 0.99 & 0.99 & 0.97 \\ 

condition\_4 & 0.97 & 0.97 & 0.94 & 0.91\\

fillers & 0.75 & 0.8 & 0.86 & 0.99  \\ 

\bottomrule
\end{tabular}%
}
\end{table}

\newpage
\section{Appendix C: Prompts versions across all experiments.}
\label{appendix:C}
\renewcommand{\arraystretch}{1.5} 
\begin{table*}[ht]
\centering
\caption{Prompts formulations across different replication versions}
\resizebox{\textwidth}{!}{%
\begin{tabular}{lp{0.8\textwidth}} 
\toprule
\textbf{Experiment} & \textbf{Prompt}\\ 
\midrule
bilinguals-zero shot & \texttt{You are a Spanish–English bilingual speaker. You are participating in a study. You will see a short sentence missing information and your task is to complete the sentence with the information appearing after it in a way that sounds as natural as possible to you. Just answer with the one word among the given options you would use to complete the sentence, don't repeat the whole sentence: '\{text\}'} \\

neologisms-zero shot & \texttt{Vous êtes de langue maternelle française. Vous participez à une étude. Indiquez si la phrase suivante contient un mot nouveau ou un mot existant employé avec un sens nouveau simplement par 'oui' ou 'non': '\{text\}' Si ce mot existe, indiquez-le sans l'expliquer}, in English: You are a native speaker of French. You are partecipating in a study. Indicate whether the following sentence contains a new word or an existing word used in a new sense with simply ‘yes’ or ‘no’: ‘\{text\}’. If there is this word, indicate it without explaining.\\

neologisms-zero shot-role & \texttt{Vous êtes de langue maternelle française. Vous participez à une étude. Indiquez si la phrase suivante contient un mot nouveau ou un mot existant employé dans un sens nouveau avec simplement 'oui' or 'non': '\{text\}' Si il y a ce mot, l'indiquez sans expliquer}, in English: You are a native speaker of French. You are partecipating in a study. Indicate whether the following sentence contains a new word or an existing word used in a new sense with simply ‘yes’ or ‘no’: ‘\{text\}’. If there is this word, indicate it without explaining.\\

neologisms-CoT & \texttt{Vous êtes de langue maternelle française. Vous participez à une étude. Tache: Indiquez si la phrase suivante contient un mot nouveau ou un mot existant employé dans un sens nouveau avec simplement 'oui' ou 'non': 'Le livre n’est distribué qu’en impadem pour le moment'. Si il y a ce mot, l'indiquez sans expliquer. Penser: Réfléchissons pas à pas. Je lis d'abord la phrase en entier et j'essaie de reconnaître tous les mots. Impadem ne me semble pas familier, alors ma réponse est: 'Oui, impadem.' Tache: Indiquez si la phrase suivante contient un mot nouveau ou un mot existant employé dans un sens nouveau avec simplement 'oui' ou 'non': 'Parler de ses angoisses aide beaucoup Valentine.' Si il y a ce mot, l'indiquez sans expliquer. Penser: Réfléchissons pas à pas. Je lis d'abord la phrase en entier et j'essaie de reconnaître tous les mots. Je les connais tous, donc ma réponse est: 'non'. Tache: Indiquez si la phrase suivante contient un mot nouveau ou un mot existant employé dans un sens nouveau avec simplement 'oui' ou 'non': '\{text\}' Si il y a ce mot, l'indiquez sans expliquer. Réponse:}, in English: You are French native speaker. You are partecipating in a study. Task: Indicate whether the following sentence contains a new word or an existing word used in a new sense with just ‘yes’ or ‘no’: ‘The book is only available in impadem at the moment’. If it does, say so without explaining. Thinking: Let's think step by step. First I read the whole sentence and try to recognise all the words. Impadem doesn't sound familiar, so my answer is: ‘Yes, impadem’. Task: Indicate whether the following sentence contains a new word or an existing word used in a new sense with just ‘yes’ or ‘no’: ‘Talking about her fears helps Valentine a lot." If it does, say so without explaining. Thinking: Let's think step by step. First I read the whole sentence and try to recognise all the words. I know them all, so my answer is ‘no’. Task: Indicate whether the following sentence contains a new word or an existing word used in a new sense with just ‘yes’ or ‘no’: : ‘\{text\}’ If there is such a word, state it without explanation. Answer:\\

\bottomrule
\end{tabular}%
}
\end{table*}

\end{document}